\documentclass{article}
\usepackage{spconf,amsmath,epsfig}

\let\OLDthebibliography\thebibliography
\renewcommand\thebibliography[1]{
  \OLDthebibliography{#1}
  \setlength{\parskip}{0pt}
  \setlength{\itemsep}{0pt plus 0.3ex}
}

\usepackage{amsmath}
\usepackage{amssymb}
\usepackage{booktabs}
\usepackage{array}
\usepackage{tabularx}
\newcolumntype{L}[1]{>{\raggedright\arraybackslash}m{#1}}
\newcolumntype{C}[1]{>{\centering\arraybackslash}p{#1}}

\usepackage{graphicx}

\begin{document}\sloppy
\topmargin=0mm

% Example definitions.
% --------------------
\def\x{{\mathbf x}}
\def\L{{\cal L}}

% Title.
% ------
\title{Human-Centered Prior-Guided and Task-Dependent Multi-Task Representation Learning for Action Recognition Pre-Training}
%
% Single address.
% ---------------
%\name{Paper ID: 521}

\name{Guanhong Wang, Keyu Lu, Yang Zhou, Zhanhao He, Gaoang Wang$^{\ast}$}

% \name{Guanhong Wang, Keyu Lu, Yang Zhou, Zhanhao He, Gaoang Wang}

%Address and e-mail should NOT be added in the submission paper. They should be present only in the camera ready paper. 
% Address.
% ---------------
% \name{First author$^{\ast}$, Second author$^{\dagger}$   (...) and Last Author$^{\ddagger}$}
% \address{$^{\ast}$First author address and e-mail; $^{\dagger}$Second author address and e-mail ; (...) and \\ $^{\ddagger}$Last author address and e-mail.}

\address{Zhejiang University-University of Illinois at Urbana-Champaign Institute, Zhejiang University, China \\guanhongwang@zju.edu.cn,
\{keyu.19, yangz.19, zhanhao.19, gaoangwang\}@intl.zju.edu.cn}

\maketitle

\renewcommand{\thefootnote}{\fnsymbol{footnote}}

\footnotetext[1]{Corresponding author.}

\begin{abstract}
Recently, much progress has been made for self-supervised action recognition. Most existing approaches emphasize the contrastive relations among videos, including appearance and motion consistency. However, two main issues remain for existing pre-training methods: 1) the learned representation is neutral and not informative for a specific task; 2) multi-task learning-based pre-training sometimes leads to sub-optimal solutions due to inconsistent domains of different tasks. To address the above issues, we propose a novel action recognition pre-training framework, which exploits human-centered prior knowledge that generates more informative representation, and avoids the conflict between multiple tasks by using task-dependent representations. Specifically, we distill knowledge from a human parsing model to enrich the semantic capability of representation. In addition, we combine knowledge distillation with contrastive learning to constitute a task-dependent multi-task framework. We achieve state-of-the-art performance on two popular benchmarks for action recognition task, i.e., UCF101 and HMDB51, verifying the effectiveness of our method.
\end{abstract}

\begin{keywords}
Multi-Task Learning, Knowledge Distillation, Video Representation Learning.
\end{keywords}

\section{Introduction}
\label{sec:intro}

% 第一段，应描述video action recognition的背景。第一步，说action recognition为什么重要，可以用于哪些方面，为什么吸引研究者去研究。第二步，介绍action recognition常用的方法有哪些。

Action recognition is a hot topic in the computer vision community. It has many practical applications such as intelligent surveillance, human-computer interaction and behavior analysis \cite{huang2018makes}. 
The critical challenge of video-based action recognition is to model the complex spatial-temporal information in videos, which is more difficult than understanding static images \cite{chen2021rspnet}.
% Early works usually either follow a two-stream paradigm \cite{simonyan2014two,feichtenhofer2017spatiotemporal,feichtenhofer2016convolutional,wang2016temporal} or exploit 3D convolutional neural network (CNN) \cite{ji20123d,tran2015learning,carreira2017quo,tran2018closer} to explore the visual appearances and temporal dynamics. 
Early works usually follow a two-stream paradigm \cite{simonyan2014two,feichtenhofer2017spatiotemporal,feichtenhofer2016convolutional,wang2016temporal} or exploit 3D convolutional neural network (3D CNN) \cite{ji20123d,tran2015learning,carreira2017quo,tran2018closer} to explore the visual appearances and temporal dynamics.
However, these supervised methods need great annotation effort \cite{wang2020self}. Recently, many works focus on action recognition pre-training with self-supervised learning frameworks without using categorical labels. These approaches usually follow the design of pretext tasks for representation learning, such as temporal shuffle \cite{lee2017unsupervised}, future frame prediction \cite{behrmann2021unsupervised}, video-based space-time cubic puzzle completion \cite{kim2019self}, and contrastive learning-based tasks \cite{han2019video,tao2020self,chen2021rspnet}.
% . are generally towards two directions. The first attempt is to design pretext tasks which can generate the supervision signals for the original video data. Such representative pretext tasks include temporal shuffle \cite{lee2017unsupervised}, future frame prediction \cite{behrmann2021unsupervised}, and video-based space-time cubic puzzle completion \cite{kim2019self}. Besides these, the other attempt is to extend the contrastive learning paradigm from images to videos, such as video dense predictive coding \cite{han2019video}, and video contrastive multi-view coding \cite{tao2020self}.

\begin{figure}[t]
    \centering
    \includegraphics[width=0.8\linewidth]{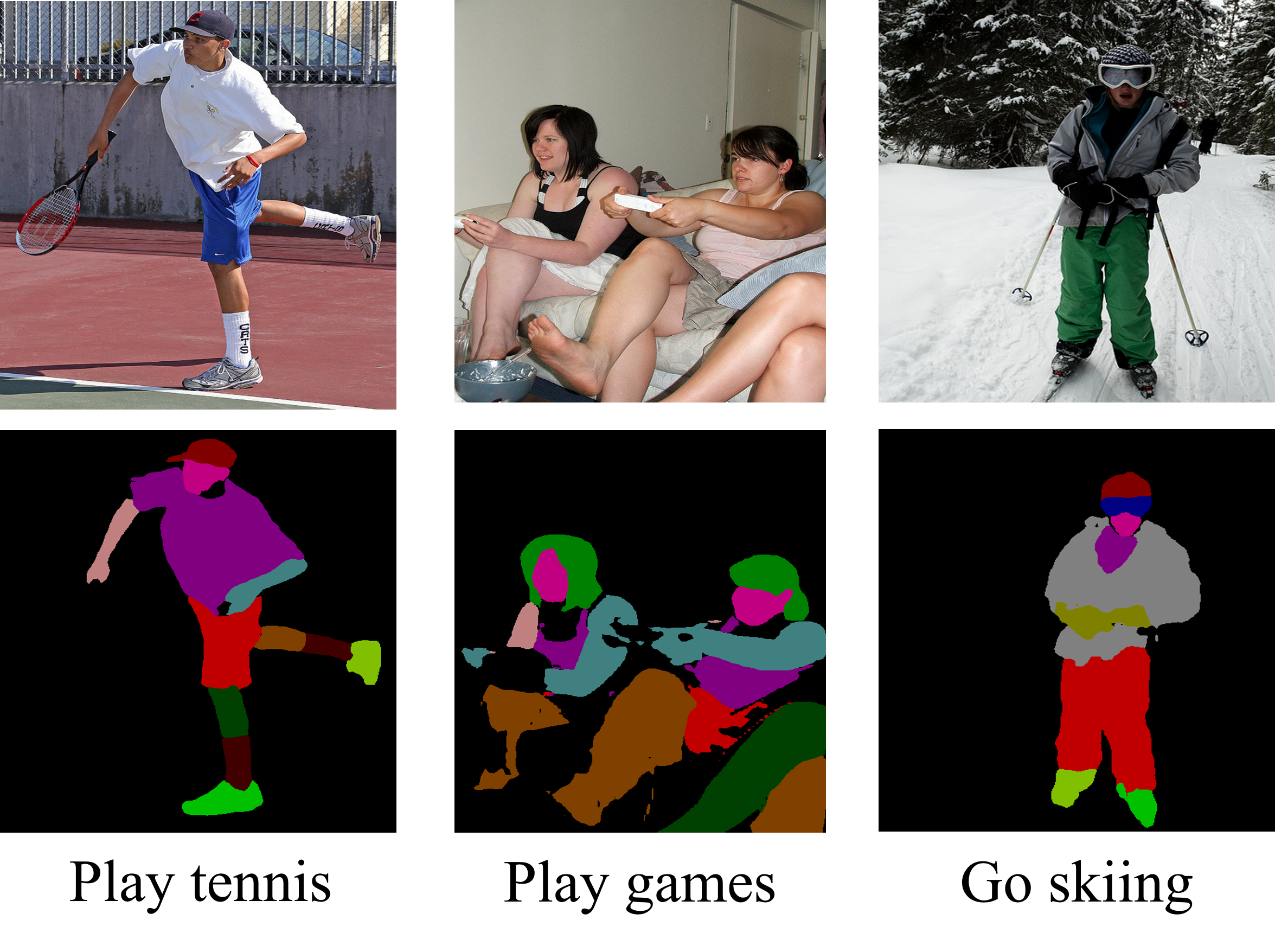}
    \vspace{-15pt}
    \caption{Two rows show some examples of human-centered actions and corresponding segmentation maps from human parsing models, respectively. The human parsing prior provides useful knowledge to action recognition tasks.}
    \label{fig:human_prior}
    \vspace{-10pt}
\end{figure}

% 第二段，介绍video action recognition的挑战，现有方法介绍，以及其不足之处。着重强调Pre-training 方法。逻辑上跟abstract对应，abstract上面写的几条，这里对每条展开描述。注意引用文献。
Although much great effort has been made for video representation learning, two main issues remain for existing pre-training methods. 
% Firstly, the learned representation by these methods is neutral and not informative for specific downstream tasks like action recognition and retrieval. This usually leads to severe performance degradation. Secondly, although some existing frameworks combine multiple pretext tasks to learn different contrastive relations among videos, the shared representation learned by the multi-task learning mechanism is usually sub-optimal due to inconsistent domains of different tasks.
% Firstly, instance discrimination based methods just pull examples sampled from the same instance (positive pairs) close to each other while repelling those from different instances (negative pairs) \cite{zhang2021incomplete}. 
The commonly used contrastive learning pre-training paradigm emphasizes the instance-level similarity \cite{zhang2021incomplete}, which can hardly capture abundant semantic information in videos, resulting in neutral and less informative representation. Therefore, it further causes severe performance degradation on many specific downstream tasks with very different objectives compared with the pre-training stage.
% , such as action recognition which concentrates on human's action classes. 
Secondly, many multi-task learning-based pre-training frameworks fail to consider the potential conflicts among different and inconsistent objectives from multiple tasks, leading to sub-optimal solutions \cite{kendall2018multi,sener2018multi}.
% Secondly, many pre-training methods exploit multiple pretext tasks to boost the performance of video representation learning by sharing parameters across tasks. However, it brings about a challenging optimization problem. Different tasks could conflict with each other due to inconsistent domains inherently. Therefore, no perfect optimal solution can optimize the performance of all tasks at the same time \cite{kendall2018multi}. Directly optimizing the average loss can be quite detrimental to a specific task’s performance \cite{liu2021conflict}. 

\begin{figure*}[t]
    \centering
    \includegraphics[width=0.85\textwidth]{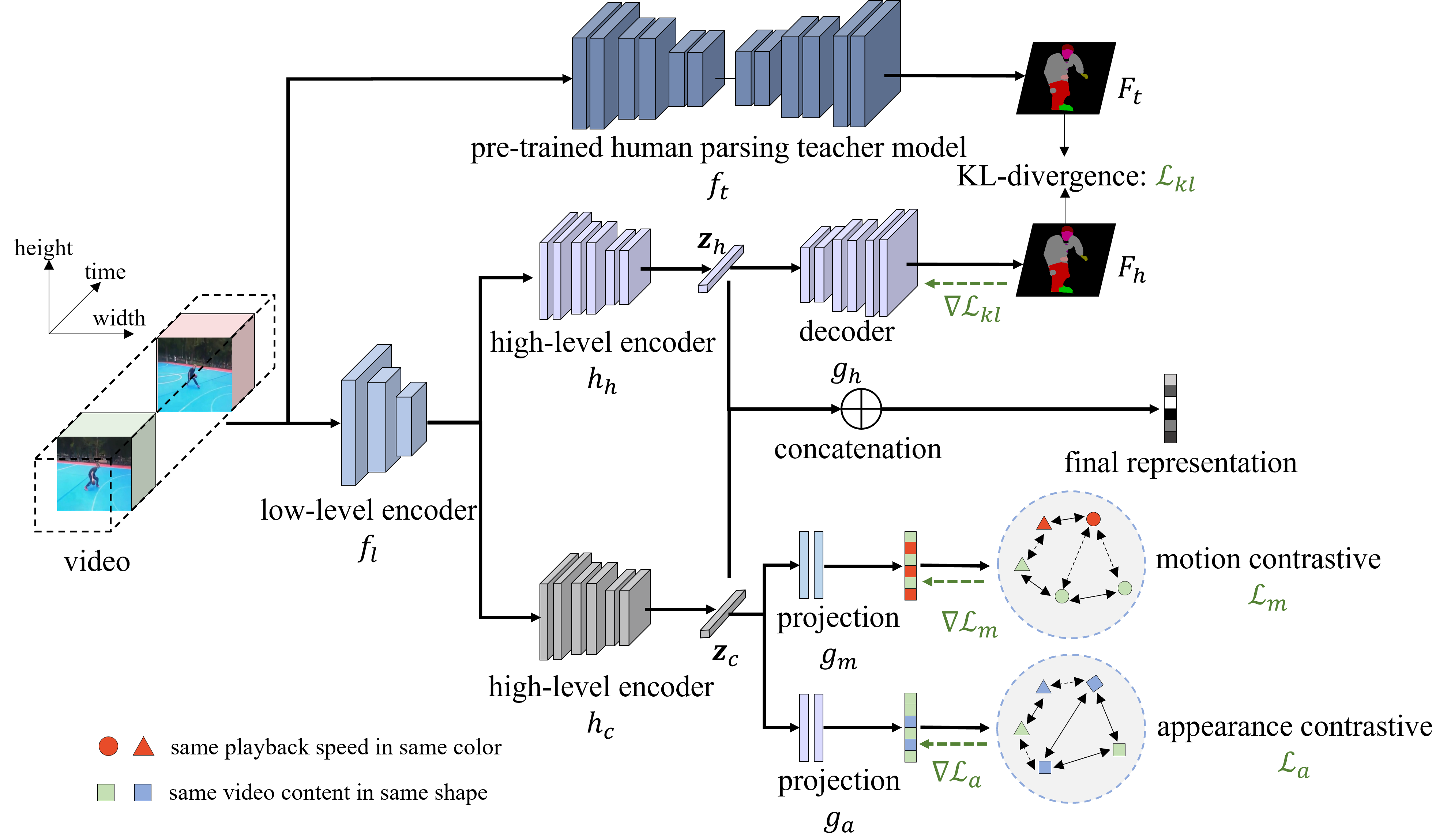}
    % 字母表示
    \vspace{-10pt}
    \caption{Overview of our proposed framework.
    % Video clips are firstly sent to a shared low-level encoder to generate common low-level feature maps across multiple tasks. 
    Video clips are firstly augmented and then sent to a shared low-level encoder to generate common low-level feature maps across multiple tasks.
    Then the low-level feature maps are sent to two distinct tasks, namely the human-centered prior knowledge distillation task and video contrastive learning task, following a multi-task learning framework. The knowledge distillation branch is supervised by a pre-trained human parsing teacher model, while the contrastive learning branch combines both motion and appearance contrastive relationships. Task-dependent embeddings are learned from two distinct high-level encoders and concatenated as the final representation. }
    \label{fig:framework}
    \vspace{-10pt}
\end{figure*}

% 第三段，介绍提出的方法，如何解决上述问题
To address the above issues, in this paper, we propose a novel prior-guided and task-dependent multi-task representation learning framework for video-based action recognition pre-training. 
% First, as shown in Fig.~\ref{fig:human_prior}, since the contents of most action recognition videos are human-centered, we incorporate the human-centered prior by distilling informative knowledge from a human parsing teacher model with an encoder-decoder network. 
First, we incorporate the human-centered prior by distilling informative knowledge from a human parsing teacher model with an encoder-decoder network. 
This is based on the intuition that the informative human parsing knowledge can reflect human actions, which is well-aligned with the downstream action recognition objectives. An example is shown in Fig.~\ref{fig:human_prior} for a better demonstration.
% The observations behind this are as follows: 1) As shown in Fig.~\ref{fig:human_prior}, since the contents of most action recognition videos are human-centered, human parsing model is good at extracting rich feature map from these videos. 2) Knowledge distillation can transfer the rich knowledge from teacher model to student model to enrich the semantic capability of the representation.
In addition, we also combine the contrastive learning with both appearance and motion consistency into a multi-task learning framework. 
%To avoid the conflict from multiple tasks and sub-optimal solutions, task-dependent representations are learned individually and further combined for downstream tasks. 
To avoid the potential conflict from multiple tasks, 
% instead of following previous works which adaptively re-weight the objectives of each task based on heuristics \cite{chen2018gradnorm,kendall2018multi}, 
task-dependent models are employed.
% we use the separate models for different tasks so that the gradients can hardly conflict with each other. 
The generated task-dependent representations are further combined for downstream tasks. The framework of our proposed method is demonstrated in Fig.~\ref{fig:framework}. 
We conduct extensive experiments for action recognition on UCF101 and HMDB51 datasets and achieve state-of-the-art (SOTA) performance, verifying the effectiveness of our proposed method. 

The main contributions of this paper are summarized as follows:
1) we present a novel framework that incorporates the human-centered prior for representation learning by knowledge distillation (KD) from the human parsing teacher model; 
2) we employ a multi-task learning framework with the task-dependent representation learning strategy; 
3) we conduct experiments on action recognition task, demonstrating the effectiveness of the proposed multi-task learning framework for video representation learning.

\section{Related Work}

\noindent \textbf{Self-Supervised Learning for Visual Representation.}
Self-supervised learning aims at learning discriminative representation by leveraging information from unlabeled data. Most works have explored the self-supervised visual representation learning based on the design of pretext tasks, such as image inpainting \cite{pathak2016context}, permutation \cite{misra2020self}, predicting jigsaw puzzles \cite{kim2018learning}, and contrastive learning strategies. 
Recently, the extension from image to video representation learning has become increasingly popular due to the richer temporal information of videos.
Like representation learning for images, the self-supervised video representation learning also focuses on the design of pretext tasks yet with the extension of considering temporal consistency in video clips. 
% contrastive learning and pretext task learning strategies. The contrastive learning usually compares both motion and appearance information among video clips. 
For example, \cite{fernando2017self,lee2017unsupervised,xu2019self} attempt to shuffle the frame and clip order along the temporal dimension; \cite{han2019video} proposes a pretext task to predict future frames; \cite{kim2019self} learns video features through designing space-time cubic puzzles, and \cite{benaim2020speednet} proposes the SpeedNet to predict the motion speed in videos. 
In addition, contrastive learning is also widely used in the design of pretext tasks. Specifically, \cite{han2019video} proposes a dense predictive coding method with contrastive loss for video frames; \cite{tao2020self} exploits contrastive multi-view video coding with the inspiration by \cite{tian2020contrastive} for image coding; \cite{chen2021rspnet} proposes a relative contrastive speed perception task to learn the motion information of videos. 
% Due to the recent success of discriminating playback speeds pretext\cite{epstein2020oops}\cite{wang2020self}\cite{yao2020video},  
% However, the learned representation by most existing self-supervised models is neutral and not much informative for a particular specific task, leading to severe performance degradation on downstream tasks.
However, the representation by self-supervised training may be not well-aligned with the objectives of downstream tasks, 
% by most existing self-supervised models lacks much informative semantic for a particular specific task, 
leading to severe performance degradation.
\vspace{+5pt}

\noindent \textbf{Video-Based Action Recognition.}
% With the rapid development of deep learning, more and more researchers are interested in video based action recognition tasks. 
% In the early years, \cite{karpathy2014large} proposed to use 2D CNN model to extract the features of each frame and fused them. However, this method does not consider the temporal dynamics of video. In order to overcome this issue, \cite{donahue2015long} started to model the temporal information by using LSTM \cite{hochreiter1997long}. 
Some works of action recognition follow a two-stream architecture to model appearance and temporal information separately. 
% The recent works usually follow two directions to improve temporal modeling ability \cite{li2020tea}. The first one follows two-stream architecture. 
For example, \cite{simonyan2014two} exploits both spatial 2D CNN and temporal CNN to extract the spatial information and motion information on RGB frames and optical flows. 
% respectively, followed by an aggregation of both features to obtain a good representation. 
% \cite{wang2016temporal} models long video clips by a sparse sampling strategy. 
% \cite{wang2016temporal} models long video clips by using a sparse sampling strategy. 
\cite{lin2019temporal} utilizes a shift module for better capturing temporal information. \cite{feichtenhofer2019slowfast} proposes a slow-fast network to extract spatial semantics and temporal motion information in the video clips. 
However, such 2D CNN-based methods adopted in the two-stream architecture have limited ability to capture the dynamics of visual tempos \cite{yang2020temporal}. 
To address such issues, some recent works of action recognition is based on 3D CNN and its variants, that extract both appearance and temporal information jointly. Specifically, \cite{tran2015learning} firstly employs 3D convolutions on adjacent frames to model the spatial and temporal information of video. \cite{carreira2017quo} inflates the pre-trained 2D convolutions to 3D convolutions kernels. \cite{tran2018closer} decomposes 3D convolution kernels into a 2D+1D paradigm to improve the performances of 3D CNNs. \cite{wang2018non} proposes a self-attention mechanism to model long-range temporal dynamics of videos. Despite the great progress has been made in recent works, how to better generate more informative representations with self-supervised pre-training framework for action recognition is still under explored and needs further study.

\section{Proposed Method}

In this section, we present our prior-guided and task-dependent multi-task representation learning framework for video-based action recognition pre-training. 
% We firstly describe how to incorporate the human-centered prior by distilling informative knowledge from a human parsing teacher model with an encoder-decoder network, to enrich the semantic capability of the representation. Then, we combine knowledge distillation (KD) with video contrastive learning into a multi-task learning framework and increase the discrimination with task-dependent representations.
We firstly describe how to incorporate the human-centered prior information by distilling informative knowledge from a human parsing teacher model with an encoder-decoder network, to enrich the semantic capability of the representation. Then, we combine knowledge distillation (KD) with video contrastive learning into a multi-task learning framework to boost the discrimination with fused task-dependent representations.

\subsection{Human-Centered Prior Knowledge Distillation}
Since most action recognition videos are related to human actions, we incorporate the human-centered prior into representation learning with a pre-trained human parsing guided teacher network. Denote $f_l(\cdot)$ and $h_{h}(\cdot)$ as the low-level encoder and high-level encoder of the human prior representation module, respectively, where the low-level encoder $f_l(\cdot)$ shares the weights across multiple tasks. Given the input video clip $\boldsymbol{X}$, the human prior representation $\boldsymbol{z}_h$ can be obtained by
\begin{equation}
    \boldsymbol{z}_h = h_h(f_l(\boldsymbol{X})).
\label{eq:z_h}
\end{equation}

To guide the learning of the human prior representation, we employ a pre-trained human parsing network $f_t(\cdot)$ as the teacher model. %Given the input video clip $\boldsymbol{X}$, the teacher model can generate the human parsing segmentation feature maps $\boldsymbol{F}_t$ as follows,
Given the middle frame of input video clip $\boldsymbol{X}_m$, the teacher model can generate the human parsing segmentation feature map $\boldsymbol{F}_t$ as follows,
% \begin{equation}
%     \boldsymbol{F}_t = f_t(\boldsymbol{X}),
% \label{eq:F}
% \end{equation}
\begin{equation}
    \boldsymbol{F}_t = f_t(\boldsymbol{X}_m),
\label{eq:F}
\end{equation}
% where $\boldsymbol{F}_t$ is already normalized with the softmax operation.
In the meanwhile, the prior representation is fed into a decoder $g_h(\cdot)$ to generate a parsing segmentation as follows,
\begin{equation}
    \boldsymbol{F}_h = g_h(\boldsymbol{z}_h).
\label{eq:F}
\end{equation}
We assume both $\boldsymbol{F}_t$ and $\boldsymbol{F}_h$ are normalized feature maps after the softmax operation. $\boldsymbol{F}_h$ is supervised by the teacher model with the KL-divergence loss $\mathcal{L}_{KL}$ as follow,
\begin{equation}
    \mathcal{L}_{KL} = -\frac{1}{N} \sum_{i,j}^N \sum_{c=1}^C \boldsymbol{F}_{t}(i,j,c) \log \boldsymbol{F}_{h}(i,j,c),
\label{eq:KL}
\end{equation}
where $i,j$ represent the spatial index, $c$ represents the segmentation class index, $C$ is the number of segmentation classes, and $N$ is the number of spatial dimensions. With human parsing guided learning, the prior representation $\boldsymbol{z}_h$ should contain rich semantic information related to human-centered actions.

\subsection{Task-Dependent Multi-Task Learning}
To learn discriminative representations among videos, inspired by \cite{chen2021rspnet}, we also combine the video contrastive learning module into a multi-task learning framework. From the output of the low-level encoder $f_l(\cdot)$, we stack another high-level encoder $h_c(\cdot)$, which is different to $h_h(\cdot)$ used in the human prior module, to generate contrastive representations $\boldsymbol{z}_c$ as follows,
\begin{equation}
    \boldsymbol{z}_c = h_c(f_l(\boldsymbol{X})).
\label{eq:z_c}
\end{equation}

Two more projection heads, \textit{i.e.}, $g_m(\cdot)$ and $g_a(\cdot)$, are further employed to learn motion contrastive and appearance contrastive with the following margin ranking loss and InfoNCE loss \cite{he2020momentum},
\begin{equation}
\begin{aligned}
    & \mathcal{L}_m = \text{max}(0, \gamma-(d^+-d^-)),\\
    & \mathcal{L}_a = -\log \frac{q^+}{q^++\sum_{n=1}^K q_n^-},
\label{eq:contrastive}
\end{aligned}
\end{equation}
where $d^{+/-}=d(g_m(\boldsymbol{z}_c),g_m(\boldsymbol{z}_c^{+/-}))$ represents the distance between the anchor motion embedding to the embedding with the same/different playback speed \cite{chen2021rspnet}, and $q^{+/-}=\exp (d(g_a(\boldsymbol{z}_c),g_a(\boldsymbol{z}_c^{+/-}))/\tau)$ represents the similarity between anchor appearance embedding and the embedding from the same/different video clips. 

With the human prior knowledge distillation module and video contrastive module, we form a multi-task learning framework with the following total loss,
\begin{equation}
    \mathcal{L} = \lambda_k \mathcal{L}_{KL}+\lambda_m \mathcal{L}_m+\lambda_a \mathcal{L}_a,
\label{eq:loss}
\end{equation}
where $\lambda_k$, $\lambda_m$ and $\lambda_a$ are the weights of individual loss. Instead of widely adopted multi-task learning approaches that different tasks share the common representation, we use task-dependent and uncorrelated representations $\boldsymbol{z}_h$ and $\boldsymbol{z}_c$ to learn different semantic information from videos, which can better avoid the conflict across different tasks. We use the concatenated representation $[\boldsymbol{z}_h || \boldsymbol{z}_c]$ as the final representation of each video clip.

\section{Experiment}

\subsection{Implementation Details}
We adopt Kinetics-400 \cite{carreira2017quo} dataset for self-supervised pre-training. We employ the SOTA human parsing model SCHP \cite{li2020self} as the pre-trained teacher model for human-centered prior knowledge distillation. After the pre-training process, we finetune the model on UCF101 \cite{soomro2012ucf101} and HMDB51 \cite{kuehne2011hmdb} datasets for downstream action recognition task. We evaluate the performance based on top-1 and top-5 accuracy (Acc@1, Acc@5, respectively). More implementation details are demonstrated in the Supplementary Material.

% \noindent \textbf{Kinetics-400}\cite{carreira2017quo} is a large scale human action recognition dataset. This dataset is devided into training/validation/testing splits. It consist of around 240,000 training videos with 400 human action classes. Each video lasts about 10 seconds. In this work, we use the training split as our pre-training dataset without any labels.

% \noindent \textbf{UCF101} \cite{soomro2012ucf101} is a widely used dataset which contains 13,320 clips with 101 action categories. It is devided into three splits. Follow the previous work\cite{xu2019self,benaim2020speednet}, we use training split 1as finetuning dataset and training/testing split 1 for evaluation. 

% \noindent \textbf{HMDB51} \cite{kuehne2011hmdb} is a dataset which consists of 6,849 clips with 51 human action classes from YouTube. It is devided into three training/testing splits. And following prior works \cite{xu2019self,benaim2020speednet}, we use training/testing split 1 for downstream task evaluation.

\begin{table}[t]
\begin{center}
\caption{Top-1 and Top-5 accuracy on UCF101 and HMDB51 datasets compared with SOTA methods. Best performance is marked in bold for each type of architecture.} 
\label{tab:tabel1}
\begin{tabular}{L{2.1cm}|C{1.05cm}C{1.05cm}|C{1.05cm}C{1.05cm}}
  \toprule
  & \multicolumn{2}{c}{UCF101} & \multicolumn{2}{c}{HMDB51}\\
  Method & Acc@1 & Acc@5 & Acc@1 & Acc@5\\
  \midrule
  \multicolumn{5}{l}{C3D Architecture}\\
  \midrule
  VCP \cite{luo2020video}  & 68.5 & - & 32.5 & -\\
  MAS \cite{wang2019self}  & 61.2 & - & 33.4 & -\\
  RTT \cite{jenni2020video}  & 69.9 & - & 39.6 & -\\
  RSPNet \cite{chen2021rspnet}   & 76.7 & - & 44.6 & - \\
  RSPNet$^*$ \cite{chen2021rspnet}   & 77.6 & 93.7 & 45.4 & 75.7 \\
  \textbf{Ours} & \textbf{80.4} & \textbf{95.7} & \textbf{46.1} & \textbf{78.4} \\
  \midrule
  \multicolumn{5}{l}{R(2+1)D Architecture}\\
  \midrule
  VCP \cite{luo2020video}  & 66.3 & - & 32.2 & -\\
  PSP \cite{cho2020self}  & 74.8 & - & 36.8 & -\\
  ClipOrder \cite{xu2019self}  & 72.4 & - & 30.9 & -\\
  PRP \cite{yao2020video}  & 72.4 & - & 35.0 & -\\
  Pace \cite{wang2020self}  & 77.1 & - & 36.6 & -\\
  RSPNet \cite{chen2021rspnet}   & 81.1 & - & 41.8  & -\\
  RSPNet$^*$ \cite{chen2021rspnet}   & 79.4 & 94.3 & 43.0 & 74.5 \\
  \textbf{Ours} & \textbf{81.6} & \textbf{95.3} & \textbf{46.1} & \textbf{74.8}\\
  \bottomrule
\end{tabular}
\end{center}
\vspace{-10pt}
\end{table}

\begin{table}[t]
\begin{center}
\caption{Ablation Studies on Action Recognition Task on UCF101 and HMDB51 datasets. Best performance is marked in bold for each type of architecture.} 
\label{tab:tabel2}
\begin{tabular}{L{2.1cm}|C{1.05cm}C{1.05cm}|C{1.05cm}C{1.05cm}}
  \toprule
  & \multicolumn{2}{c}{UCF101} & \multicolumn{2}{c}{HMDB51}\\
  Method & Acc@1 & Acc@5 & Acc@1 & Acc@5\\
  \midrule
  \multicolumn{5}{l}{C3D Architecture}\\
  \midrule
  w/o KD    & 77.6 & 93.7 & 45.4 & 75.7 \\
  TI    & 79.3 & 92.1  & 44.8   & 67.5  \\
  \textbf{Full model} & \textbf{80.4} & \textbf{95.7} & \textbf{46.1} & \textbf{78.4} \\
  \midrule
  \multicolumn{5}{l}{R(2+1)D Architecture}\\
  \midrule
  w/o KD    & 79.4 & 94.3 & 43.0  & 74.5\\
  TI    & 80.1 & 94.3 & 44.6 & 74.4 \\
  \textbf{Full model} & \textbf{81.6} & \textbf{95.3} & \textbf{46.1} & \textbf{74.8}\\
  \bottomrule
\end{tabular}
\end{center}
\vspace{-10pt}
\end{table}

\begin{figure}[t]
    \centering
    \includegraphics[width=0.95\linewidth]{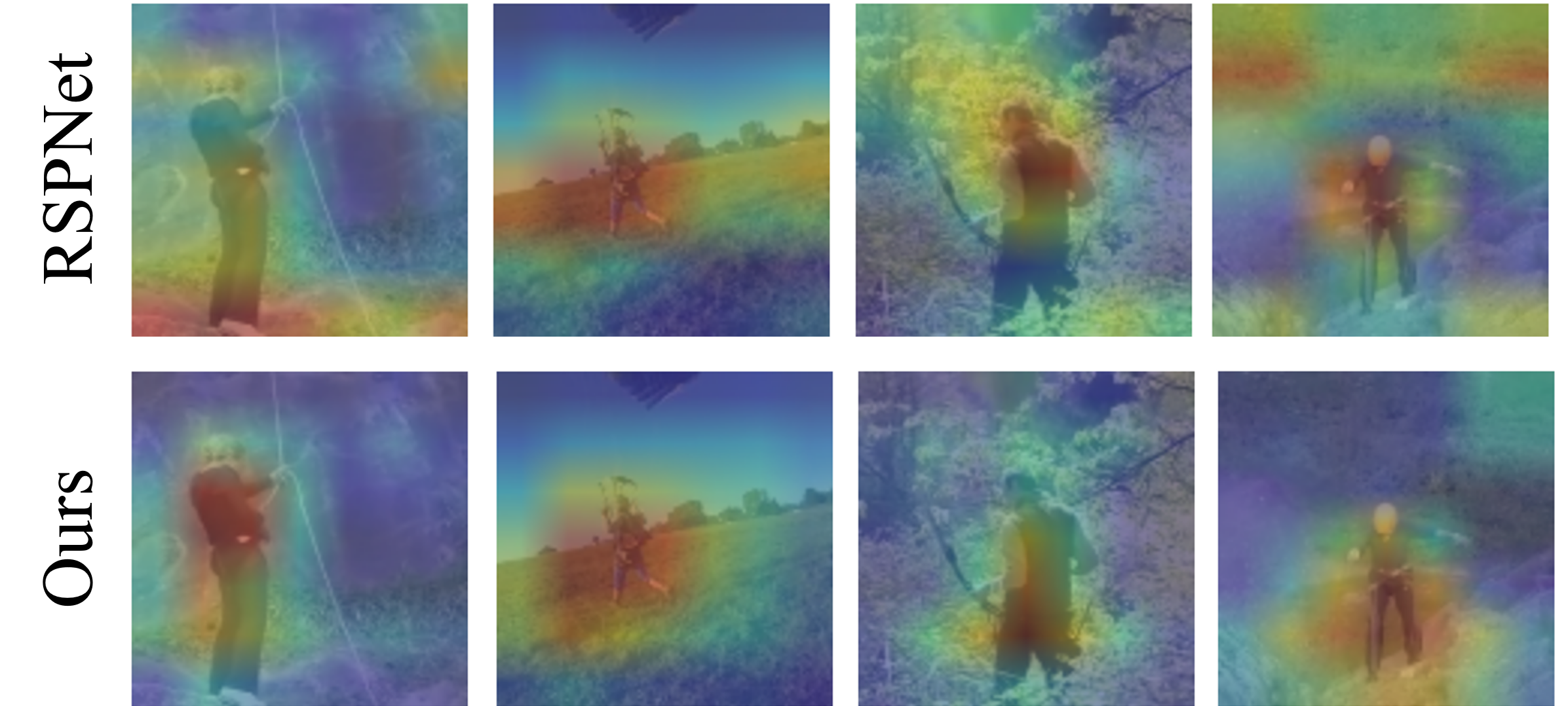}
    \vspace{-10pt}
    \caption{Some examples of visualization results. The first row and second row represent RSPNet \cite{chen2021rspnet} and our proposed method, respectively. As expected, our model focuses more on the human actions rather than background regions. }
    \label{fig:visualization}
    \vspace{-10pt}
\end{figure}

\subsection{Evaluation on Video Action Recognition Task}

\noindent \textbf{Comparison with SOTA Methods.} We compare our method with other SOTA methods on UCF101 and HMDB51 datasets. We report top-1 and top-5 accuracy with C3D \cite{tran2015learning} and R(2+1)D \cite{tran2018closer} architectures in Table~\ref{tab:tabel1}. RSPNet$^*$ denotes that a 100-epoch finetuning is used instead of the result in the original paper.
It is shown that we outperform all the other SOTA methods with a large margin, verifying the effectiveness of our proposed method. 
% Except for a slightly worse result compared with RSPNet$^*$ with C3D on HMDB51, we achieve the best performance on the all the other settings with a large margin, verifying the effectiveness of our proposed method.  
% From the Table, we can see that our method achieve the state-of-the-art results on UCF101. To be specific, our method outperforms RTT by 10.5\% with the C3D backbone. And with R(2+1)D backbone, compared with Pace, our method improves the accuracy from 77.0\% to 81.6\%. Moveover, we compare our method with RSPNet wich exploits both relative playback speed and instance discrimination as pretexts. Our method improve the performance from 77.6\% to 80.4\% with C3D backbone and from 79.4\% to 81.6\% with R(2+1)D backbone. Although our method is slight lower than RSPNet with C3D network on HMDB51 dataset, our method outperforms it by 3.1\%. This is because that HMDB51 is smaller and less informative than UCF101.
\vspace{+5pt}

\noindent \textbf{Ablation Study.} 
% Furthermore, we conduct ablation studies on the effectiveness of each individual components of our proposed method. 
We conduct ablation studies on the effectiveness of each component of our proposed method on UCF101 and HMDB51 datasets with the results shown in Table~\ref{tab:tabel2}.
% The experiments are conducted on both C3D and R(2+1)D architectures on UCF101 and HMDB51 datasets. 
% We report Acc@1 and Acc@5 results in Table~\ref{tab:tabel2}. 
Denote the removal of the knowledge distillation module as ``w/o KD", and the use of task-independent representation with the shared representation across multiple tasks as ``TI". It can be found that there is a significant degradation for the corresponding variants compared with the full model, further demonstrating the effectiveness of our proposed method on individual components.
\vspace{+5pt}
% In Table \ref{tab:tabel2}, we present the top-1 and top-5 accuracy results of knowledge distillation on UCF101. It can be found that human-centered knowledge distillation is helpful for improving the performance of action recognition with C3D backbone or R(2+1)D backbone.

% \subsection{Evaluation on Video Retrieval Task}
% We evaluate the proposed algorithm against state-of-the-art methods on video retrieval benchmarks. Given a query video with feature representation being, we use the nearest neighbor search to retrieval relevant videos based on the cosine similarity. We evaluate our metod on split 1 of the UCF101 dataset and apply the top-k accuracies as evaluation metrics.

\noindent \textbf{Visualization.}
% For better understanding the effectiveness of our framework, we show some visualization examples of the region of interest by using class-activation map (CAM) technique \cite{zhou2016learning} in Fig.~\ref{fig:visualization}. 
To better reveal the effectiveness of our framework, we show some visualization examples of the region of interest by using class-activation map (CAM) technique \cite{zhou2016learning} in Fig.~\ref{fig:visualization}. 
The produced heatmap is added on the original video frames for each example in the figure. The first row shows the results from RSPNet \cite{chen2021rspnet}, while the second row represents the results of our proposed method.
% Fig.~ \ref{fig:visualization} shows the heatmaps of video clips of our method compared baseline RSP \cite{chen2021rspnet}. In the experiment setting, the middle frame of video is adopted to visualize the heatmap to represent the learning ability of appearance information. 
% It is shown that our method concentrates more on the human body, while RSPNet is distracted by uncorrelated background regions.
It is shown that our method concentrates more on the human action, while RSPNet is distracted by uncorrelated background regions.
% The visualization results demonstrate the validity of our motivation that borrows human-centered prior for multi-task pre-training for learning discriminativ e representations.
The visualization results demonstrate the validity of our motivation that incorporates human-centered prior knowledge for multi-task pre-training to learn discriminative representations.
More visualization examples are included in Supplementary Material.

\section{Conclusion}
% In this paper, we present a novel framework that incorporates the human-centered prior for representation learning by knowledge distillation from human parsing teacher model. Moreover, we employ the multi-task learning framework with the task-dependent representation learning strategy. Extensive experiments are conducted on action recognition task, demonstrating the effectiveness of the proposed multi-task learning framework for video representation learning.

In this paper, we present a novel prior-guided and task-dependent multi-task representation learning framework for video-based action recognition pre-training. First, we incorporate the human-centered prior information by knowledge distillation from the human parsing teacher model to enrich the learned representation. Moreover, we follow the multi-task learning framework with the task-dependent representation learning strategy to solve the conflict of multi-task training paradigm. Experimental results on the action recognition task demonstrate the effectiveness of the proposed multi-task learning framework for video representation learning.

% \medskip
\noindent \textbf{Acknowledgement}. This work was supported by the National Natural Science Foundation of China under Grant 62106219.

%
% \begin{figure}[t]
% \begin{minipage}[b]{1.0\linewidth}
%   \centering
% % \centerline{\epsfig{figure=image1.ps,width=8.5cm}}
%   \vspace{1.5cm}
%   \centerline{(a) Result 1}\medskip
% \end{minipage}
% %
% \begin{minipage}[b]{.48\linewidth}
%   \centering
% % \centerline{\epsfig{figure=image3.ps,width=4.0cm}}
%   \vspace{1.5cm}
%   \centerline{(b) Results 2}\medskip
% \end{minipage}
% \hfill
% \begin{minipage}[b]{0.48\linewidth}
%   \centering
% % \centerline{\epsfig{figure=image4.ps,width=4.0cm}}
%   \vspace{1.5cm}
%   \centerline{(c) Result 3}\medskip
% \end{minipage}
% %
% \caption{Example of placing a figure with experimental results.}
% \label{fig:res}
% \end{figure}

\bibliographystyle{IEEEbib}
\bibliography{icme2022template}

\begin{thebibliography}{10}

\bibitem{huang2018makes}
De-An Huang, Vignesh Ramanathan, Dhruv Mahajan, Lorenzo Torresani, Manohar
  Paluri, Li~Fei-Fei, and Juan~Carlos Niebles,
\newblock ``What makes a video a video: Analyzing temporal information in video
  understanding models and datasets,''
\newblock in {\em Proceedings of the IEEE Conference on Computer Vision and
  Pattern Recognition}, 2018, pp. 7366--7375.

\bibitem{chen2021rspnet}
Peihao Chen, Deng Huang, Dongliang He, Xiang Long, Runhao Zeng, Shilei Wen,
  Mingkui Tan, and Chuang Gan,
\newblock ``Rspnet: Relative speed perception for unsupervised video
  representation learning,''
\newblock in {\em Proceedings of the AAAI Conference on Artificial
  Intelligence}, 2021, vol.~35, pp. 1045--1053.

\bibitem{simonyan2014two}
Karen Simonyan and Andrew Zisserman,
\newblock ``Two-stream convolutional networks for action recognition in
  videos,''
\newblock in {\em Advances in Neural Information Processing Systems}, 2014, pp.
  568--576.

\bibitem{feichtenhofer2017spatiotemporal}
Christoph Feichtenhofer, Axel Pinz, and Richard~P Wildes,
\newblock ``Spatiotemporal multiplier networks for video action recognition,''
\newblock in {\em Proceedings of the IEEE Conference on Computer Vision and
  Pattern Recognition}, 2017, pp. 4768--4777.

\bibitem{feichtenhofer2016convolutional}
Christoph Feichtenhofer, Axel Pinz, and Andrew Zisserman,
\newblock ``Convolutional two-stream network fusion for video action
  recognition,''
\newblock in {\em Proceedings of the IEEE Conference on Computer Vision and
  Pattern Recognition}, 2016, pp. 1933--1941.

\bibitem{wang2016temporal}
Limin Wang, Yuanjun Xiong, Zhe Wang, Yu~Qiao, Dahua Lin, Xiaoou Tang, and Luc
  Van~Gool,
\newblock ``Temporal segment networks: Towards good practices for deep action
  recognition,''
\newblock in {\em European Conference on Computer Vision}. Springer, 2016, pp.
  20--36.

\bibitem{ji20123d}
Shuiwang Ji, Wei Xu, Ming Yang, and Kai Yu,
\newblock ``3d convolutional neural networks for human action recognition,''
\newblock {\em IEEE Transactions on Pattern Analysis and Machine Intelligence},
  vol. 35, no. 1, pp. 221--231, 2012.

\bibitem{tran2015learning}
Du~Tran, Lubomir Bourdev, Rob Fergus, Lorenzo Torresani, and Manohar Paluri,
\newblock ``Learning spatiotemporal features with 3d convolutional networks,''
\newblock in {\em Proceedings of the IEEE International Conference on Computer
  Vision}, 2015, pp. 4489--4497.

\bibitem{carreira2017quo}
Joao Carreira and Andrew Zisserman,
\newblock ``Quo vadis, action recognition? a new model and the kinetics
  dataset,''
\newblock in {\em Proceedings of the IEEE Conference on Computer Vision and
  Pattern Recognition}, 2017, pp. 6299--6308.

\bibitem{tran2018closer}
Du~Tran, Heng Wang, Lorenzo Torresani, Jamie Ray, Yann LeCun, and Manohar
  Paluri,
\newblock ``A closer look at spatiotemporal convolutions for action
  recognition,''
\newblock in {\em Proceedings of the IEEE Conference on Computer Vision and
  Pattern Recognition}, 2018, pp. 6450--6459.

\bibitem{wang2020self}
Jiangliu Wang, Jianbo Jiao, and Yun-Hui Liu,
\newblock ``Self-supervised video representation learning by pace prediction,''
\newblock in {\em European Conference on Computer Vision}. Springer, 2020, pp.
  504--521.

\bibitem{lee2017unsupervised}
Hsin-Ying Lee, Jia-Bin Huang, Maneesh Singh, and Ming-Hsuan Yang,
\newblock ``Unsupervised representation learning by sorting sequences,''
\newblock in {\em Proceedings of the IEEE International Conference on Computer
  Vision}, 2017, pp. 667--676.

\bibitem{behrmann2021unsupervised}
Nadine Behrmann, Jurgen Gall, and Mehdi Noroozi,
\newblock ``Unsupervised video representation learning by bidirectional feature
  prediction,''
\newblock in {\em Proceedings of the IEEE Winter Conference on Applications of
  Computer Vision}, 2021, pp. 1670--1679.

\bibitem{kim2019self}
Dahun Kim, Donghyeon Cho, and In~So Kweon,
\newblock ``Self-supervised video representation learning with space-time cubic
  puzzles,''
\newblock in {\em Proceedings of the AAAI Conference on Artificial
  Intelligence}, 2019, vol.~33, pp. 8545--8552.

\bibitem{han2019video}
Tengda Han, Weidi Xie, and Andrew Zisserman,
\newblock ``Video representation learning by dense predictive coding,''
\newblock in {\em Proceedings of the IEEE International Conference on Computer
  Vision Workshops}, 2019, pp. 0--0.

\bibitem{tao2020self}
Li~Tao, Xueting Wang, and Toshihiko Yamasaki,
\newblock ``Self-supervised video representation learning using inter-intra
  contrastive framework,''
\newblock in {\em Proceedings of the 28th ACM International Conference on
  Multimedia}, 2020, pp. 2193--2201.

\bibitem{zhang2021incomplete}
Lin Zhang, Qi~She, Zhengyang Shen, and Changhu Wang,
\newblock ``How incomplete is contrastive learning? an inter-intra variant dual
  representation method for self-supervised video recognition,''
\newblock {\em arXiv e-prints}, pp. arXiv--2107, 2021.

\bibitem{kendall2018multi}
Alex Kendall, Yarin Gal, and Roberto Cipolla,
\newblock ``Multi-task learning using uncertainty to weigh losses for scene
  geometry and semantics,''
\newblock in {\em Proceedings of the IEEE Conference on Computer Vision and
  Pattern Recognition}, 2018, pp. 7482--7491.

\bibitem{sener2018multi}
Ozan Sener and Vladlen Koltun,
\newblock ``Multi-task learning as multi-objective optimization,''
\newblock in {\em Proceedings of the 32nd International Conference on Neural
  Information Processing Systems}, 2018, pp. 525--536.

\bibitem{pathak2016context}
Deepak Pathak, Philipp Krahenbuhl, Jeff Donahue, Trevor Darrell, and Alexei~A
  Efros,
\newblock ``Context encoders: Feature learning by inpainting,''
\newblock in {\em Proceedings of the IEEE Conference on Computer Vision and
  Pattern Recognition}, 2016, pp. 2536--2544.

\bibitem{misra2020self}
Ishan Misra and Laurens van~der Maaten,
\newblock ``Self-supervised learning of pretext-invariant representations,''
\newblock in {\em Proceedings of the IEEE Conference on Computer Vision and
  Pattern Recognition}, 2020, pp. 6707--6717.

\bibitem{kim2018learning}
Dahun Kim, Donghyeon Cho, Donggeun Yoo, and In~So Kweon,
\newblock ``Learning image representations by completing damaged jigsaw
  puzzles,''
\newblock in {\em 2018 IEEE Winter Conference on Applications of Computer
  Vision}. IEEE, 2018, pp. 793--802.

\bibitem{fernando2017self}
Basura Fernando, Hakan Bilen, Efstratios Gavves, and Stephen Gould,
\newblock ``Self-supervised video representation learning with odd-one-out
  networks,''
\newblock in {\em Proceedings of the IEEE Conference on Computer Vision and
  Pattern Recognition}, 2017, pp. 3636--3645.

\bibitem{xu2019self}
Dejing Xu, Jun Xiao, Zhou Zhao, Jian Shao, Di~Xie, and Yueting Zhuang,
\newblock ``Self-supervised spatiotemporal learning via video clip order
  prediction,''
\newblock in {\em Proceedings of the IEEE Conference on Computer Vision and
  Pattern Recognition}, 2019, pp. 10334--10343.

\bibitem{benaim2020speednet}
Sagie Benaim, Ariel Ephrat, Oran Lang, Inbar Mosseri, William~T Freeman,
  Michael Rubinstein, Michal Irani, and Tali Dekel,
\newblock ``Speednet: Learning the speediness in videos,''
\newblock in {\em Proceedings of the IEEE Conference on Computer Vision and
  Pattern Recognition}, 2020, pp. 9922--9931.

\bibitem{tian2020contrastive}
Yonglong Tian, Dilip Krishnan, and Phillip Isola,
\newblock ``Contrastive multiview coding,''
\newblock in {\em European Conference on Computer Vision}. Springer, 2020, pp.
  776--794.

\bibitem{lin2019temporal}
Ji~Lin, Chuang Gan, and Song Han,
\newblock ``Temporal shift module for efficient video understanding. 2019
  ieee,''
\newblock in {\em Proceedings of the IEEE International Conference on Computer
  Vision}, 2019, pp. 7082--7092.

\bibitem{feichtenhofer2019slowfast}
Christoph Feichtenhofer, Haoqi Fan, Jitendra Malik, and Kaiming He,
\newblock ``Slowfast networks for video recognition,''
\newblock in {\em Proceedings of the IEEE International Conference on Computer
  Vision}, 2019, pp. 6202--6211.

\bibitem{yang2020temporal}
Ceyuan Yang, Yinghao Xu, Jianping Shi, Bo~Dai, and Bolei Zhou,
\newblock ``Temporal pyramid network for action recognition,''
\newblock in {\em Proceedings of the IEEE Conference on Computer Vision and
  Pattern Recognition}, 2020, pp. 591--600.

\bibitem{wang2018non}
Xiaolong Wang, Ross Girshick, Abhinav Gupta, and Kaiming He,
\newblock ``Non-local neural networks,''
\newblock in {\em Proceedings of the IEEE Conference on Computer Vision and
  Pattern Recognition}, 2018, pp. 7794--7803.

\bibitem{he2020momentum}
Kaiming He, Haoqi Fan, Yuxin Wu, Saining Xie, and Ross Girshick,
\newblock ``Momentum contrast for unsupervised visual representation
  learning,''
\newblock in {\em Proceedings of the IEEE Conference on Computer Vision and
  Pattern Recognition}, 2020, pp. 9729--9738.

\bibitem{li2020self}
Peike Li, Yunqiu Xu, Yunchao Wei, and Yi~Yang,
\newblock ``Self-correction for human parsing,''
\newblock {\em IEEE Transactions on Pattern Analysis and Machine Intelligence},
  , no. 01, pp. 1--1, 2020.

\bibitem{soomro2012ucf101}
Khurram Soomro, Amir~Roshan Zamir, and Mubarak Shah,
\newblock ``Ucf101: A dataset of 101 human actions classes from videos in the
  wild,''
\newblock {\em arXiv preprint arXiv:1212.0402}, 2012.

\bibitem{kuehne2011hmdb}
Hildegard Kuehne, Hueihan Jhuang, Est{\'\i}baliz Garrote, Tomaso Poggio, and
  Thomas Serre,
\newblock ``Hmdb: a large video database for human motion recognition,''
\newblock in {\em Proceedings of the IEEE International Conference on Computer
  Vision}. IEEE, 2011, pp. 2556--2563.

\bibitem{luo2020video}
Dezhao Luo, Chang Liu, Yu~Zhou, Dongbao Yang, Can Ma, Qixiang Ye, and Weiping
  Wang,
\newblock ``Video cloze procedure for self-supervised spatio-temporal
  learning,''
\newblock in {\em Proceedings of the AAAI Conference on Artificial
  Intelligence}, 2020, vol.~34, pp. 11701--11708.

\bibitem{wang2019self}
Jiangliu Wang, Jianbo Jiao, Linchao Bao, Shengfeng He, Yunhui Liu, and Wei Liu,
\newblock ``Self-supervised spatio-temporal representation learning for videos
  by predicting motion and appearance statistics,''
\newblock in {\em Proceedings of the IEEE Conference on Computer Vision and
  Pattern Recognition}, 2019, pp. 4006--4015.

\bibitem{jenni2020video}
Simon Jenni, Givi Meishvili, and Paolo Favaro,
\newblock ``Video representation learning by recognizing temporal
  transformations,''
\newblock in {\em European Conference on Computer Vision}. Springer, 2020, pp.
  425--442.

\bibitem{cho2020self}
Hyeon Cho, Taehoon Kim, Hyung~Jin Chang, and Wonjun Hwang,
\newblock ``Self-supervised spatio-temporal representation learning using
  variable playback speed prediction,''
\newblock {\em arXiv preprint arXiv:2003.02692}, vol. 3, no. 6, pp. 7, 2020.

\bibitem{yao2020video}
Yuan Yao, Chang Liu, Dezhao Luo, Yu~Zhou, and Qixiang Ye,
\newblock ``Video playback rate perception for self-supervised spatio-temporal
  representation learning,''
\newblock in {\em Proceedings of the IEEE Conference on Computer Vision and
  Pattern Recognition}, 2020, pp. 6548--6557.

\bibitem{zhou2016learning}
Bolei Zhou, Aditya Khosla, Agata Lapedriza, Aude Oliva, and Antonio Torralba,
\newblock ``Learning deep features for discriminative localization,''
\newblock in {\em Proceedings of the IEEE Conference on Computer Vision and
  Pattern Recognition}, 2016, pp. 2921--2929.

\end{thebibliography}
\end{document}